\documentclass[letterpaper, 10 pt, conference]{ieeeconf}  

\IEEEoverridecommandlockouts                              

\usepackage{graphics} 
\usepackage{amsmath} 
\usepackage{amssymb}  
\usepackage{bbm}
\usepackage{algorithm}
\usepackage{algorithmic}
\usepackage{graphicx}

\usepackage{booktabs}
\usepackage{colortbl}
\usepackage{xcolor}
\usepackage{multicol}
\usepackage{multirow}

\title{\LARGE \bf
PLAT: Sparse Timed Keyframe Motion Tracking for Humanoid Control via Privileged Latent Transition Learning
}

\author{Zepeng Wang$^{1,2}$, Jiangxing Wang$^{2}$, Chao Ma$^{1}$, Xiaochuan Shi$^{1,*}$ and Zongqing Lu$^{2,3}$
\thanks{$^{1}$Wuhan University. 
$^{2}$BeingBeyond.
$^{3}$Peking University. 
}
\thanks{
$^{*}$Corresponding Author: \texttt{shixiaochuan@whu.edu.cn}.
}
}

\begin{document}

\maketitle
\thispagestyle{empty}
\pagestyle{empty}


\begin{abstract}

Humanoid motion tracking policies rely on dense frame-by-frame references, limiting their use as high-level motion controllers for planning and interactive motion generation. We study \emph{Sparse Timed Keyframe Motion Tracking}, where a policy receives only sparse future keyframes and their desired arrival times, and must execute stable whole-body motions that reach successive goals.
We propose \textbf{PLAT}, a three-stage sparse timed keyframe motion tracking policy learning framework with \textbf{P}rivileged \textbf{LA}tent \textbf{T}ransition learning. PLAT bridges dense motion tracking and sparse goal-conditioned control by exploiting dense goal sequences as privileged supervision during training while requiring only sparse timed keyframe commands at deployment. A pretrained dense tracking expert first provides robust motion priors. A privileged latent prior is then learned through DAgger-style imitation, followed by latent residual reinforcement learning that refines latent transitions instead of directly optimizing actions.
Extensive simulation experiments demonstrate that PLAT maintains accurate and stable sparse timed keyframe tracking across varying planning horizons, with particularly strong performance under long-horizon commands. Successful deployment on a Unitree G1 humanoid robot further demonstrates the effectiveness and practicality of PLAT for sparse humanoid motion control.

\end{abstract}


\section{INTRODUCTION}

Recent advances in reinforcement learning and large-scale motion imitation have significantly improved humanoid whole-body control \cite{luo2025sonic,zeng2026scaling,qi2026humanoid}. By tracking dense frame-by-frame reference trajectories, modern motion tracking policies can reproduce agile locomotion, dynamic acrobatic skills, and complex whole-body behaviors with impressive robustness \cite{wang2026omnixtreme,sun2026mosaic,zhang2025track}. Benefiting from increasingly diverse motion datasets and scalable training pipelines, dense motion tracking has become one of the dominant paradigms for general-purpose humanoid control and has demonstrated successful transfer to real humanoid robots.

Despite these advances, dense motion tracking fundamentally relies on continuous reference supervision. During execution, a reference pose is provided at every control step, allowing the policy to focus primarily on local tracking rather than long-horizon decision making. While this formulation is highly effective for motion imitation, it provides limited flexibility for downstream applications. In practical scenarios such as interactive motion editing, task-level planning, teleoperation, or language-driven motion generation, users naturally specify only sparse goals or key poses instead of dense trajectories. Converting these sparse commands into dense reference motions typically requires additional planning or motion generation modules \cite{tevet2025closd,rempe2026kimodo,wang2026motionbricks}, resulting in a multi-stage pipeline that is difficult to optimize jointly.

These limitations motivate a new control paradigm, \textit{Sparse Timed Keyframe Motion Tracking}, in which a humanoid policy receives only sparse future keyframes together with their desired arrival times.
Rather than following dense frame-by-frame references, the policy must infer appropriate whole-body transitions that accurately reach successive keyframes while preserving balance and motion stability.
Compared with conventional dense tracking, sparse timed commands specify only the desired destination of a motion, leaving the intermediate transition implicit.
This incomplete specification becomes increasingly challenging as the planning horizon grows, where the absence of dense reference correction makes errors more likely to accumulate over time.

A straightforward solution is to train a goal-conditioned policy directly with sparse reinforcement learning. However, sparse supervision leads to difficult exploration and often produces unstable behaviors. Another approach is to distill a dense tracking expert through DAgger using sparse observations. While online imitation alleviates distribution shift, the policy still learns to predict expert actions directly from under-specified sparse commands, without explicitly modeling the intermediate transition implied by the dense reference motion. As the planning horizon increases, this missing transition information makes accurate and stable sparse tracking increasingly difficult.

Our key observation is that dense motion tracking already contains rich information about how humans naturally transition between distant poses. Although dense reference sequences are unavailable during deployment, they can serve as privileged supervision during training to learn a compact latent representation of motion transitions. Once such a transition prior is learned, sparse keyframe control can be formulated as predicting appropriate latent transitions instead of directly searching for actions under sparse rewards.

Motivated by this insight, we propose \textbf{PLAT}, a three-stage sparse timed keyframe motion tracking policy learning framework with \textbf{P}rivileged \textbf{LA}tent \textbf{T}ransition learning. PLAT bridges dense motion tracking and sparse goal-conditioned control through latent transition learning. A dense motion tracking expert is first pretrained to provide robust motion priors. Dense future goal sequences are then exploited as privileged supervision to learn a goal-conditioned latent transition prior via DAgger-style imitation, while only sparse timed keyframe commands are required during deployment. Finally, we introduce a latent residual reinforcement learning stage that predicts residual corrections in the latent space rather than directly optimizing actions, preserving the learned motion prior while improving timed keyframe accuracy and long-horizon robustness.

Our main contributions are summarized as follows:

\begin{enumerate}

\item We formulate \textit{Sparse Timed Keyframe Motion Tracking}, a new humanoid control task in which a policy reaches sparse future keyframes at commanded horizons while maintaining stable long-horizon whole-body motion.
\item We propose PLAT, a three-stage framework that bridges dense motion tracking and sparse goal-conditioned control through privileged latent transition learning and latent residual reinforcement learning.
\item We conduct extensive simulation and real-world experiments demonstrating that PLAT achieves accurate and stable sparse timed keyframe tracking across varying planning horizons, with particularly strong performance under long-horizon commands, and transfers successfully to a Unitree G1 humanoid robot.
\end{enumerate}
\section{RELATED WORK}

\subsection{Dense Motion Tracking for Humanoid Control}

Reference-conditioned motion tracking has become a central mechanism for learning expressive whole-body humanoid control from motion capture, retargeted animation, and generated motion data.
Recent simulated-avatar and robot systems have scaled tracking policies to large motion libraries, noisy perceptual inputs, and real-world humanoid deployment, including perpetual avatar control \cite{luo2023phc}, universal humanoid motion representations \cite{luo2024uhm}, real-time human-to-humanoid teleoperation \cite{he2024h2o,he2024omnih2o,fu2024humanplus}, and sim-to-real refinement for agile whole-body skills \cite{he2025asap,xie2025kungfubot}.
Among them, SONIC~\cite{luo2025sonic} explores large-scale humanoid motion tracking by scaling model capacity, dataset volume, and training compute, achieving highly natural and robust whole-body tracking.

Despite their impressive tracking performance, these methods assume dense frame-by-frame reference motions throughout execution.
In contrast, our work studies sparse timed keyframe tracking, where dense trajectories are exploited only during training while the deployed policy relies solely on sparse timed commands.

\subsection{Sparse Motion Control}

Sparse keyframe-conditioned control provides a compact command interface by specifying a small number of desired future states, poses, or partial constraints rather than a complete frame-by-frame reference trajectory.
Recent work has studied sparse inputs for physics-based retargeting~\cite{reda2023sparseinputs}, timed keyframe sequences for legged locomotion~\cite{zargarbashi2024robotkeyframing}, and unified character controllers that can respond to masked keyframes, text, and scene conditions~\cite{tessler2024maskedmimic}.
More recently, BFM-Zero~\cite{li2025bfm} unifies multiple humanoid control tasks, including pose reaching, reward optimization and motion tracking through a behavior foundation model, while AnyBody~\cite{li2026anybody} further explores a unified whole-body humanoid controller driven by arbitrary subsets of body keypoints.
These works collectively demonstrate the growing trend toward sparse, high-level motion specifications as a practical interface for humanoid control.

However, existing methods primarily formulate sparse control as goal reaching, motion completion, or behavior generation, without explicitly considering timed keyframe tracking.
Our formulation requires the policy to reach future keyframes at commanded horizons while maintaining stable whole-body motions, introducing substantial transition ambiguity that is absent in conventional dense tracking.

\subsection{Privileged Latent Representation Learning}

Latent motion priors and sequence encoders are widely used to represent multimodal transitions, reusable skills, and compact control spaces \cite{li2026coordex}.
Recent humanoid and character-control methods have learned variational or hybrid latent spaces from motion-tracking policies \cite{luo2024uhm,yin2025unitracker}, combined discrete and continuous latent representations for sparse-goal control and motion in-betweening \cite{bae2025hybridlatent}, or distilled tracking controllers into structured latent priors \cite{tan2026slmp}.
Masked motion inpainting also shows that training with partial motion descriptions can produce flexible control behavior across different conditioning modalities \cite{tessler2024maskedmimic}.

Different from these methods, our privileged information is not future robot observations or environment states, but the dense intermediate transition between sparse timed keyframes.
PLAT therefore learns a deployable latent transition prior that explicitly models motion evolution under sparse timed commands.
\section{Method}

\begin{figure*}[t]
    \centering
    \includegraphics[width=\textwidth]{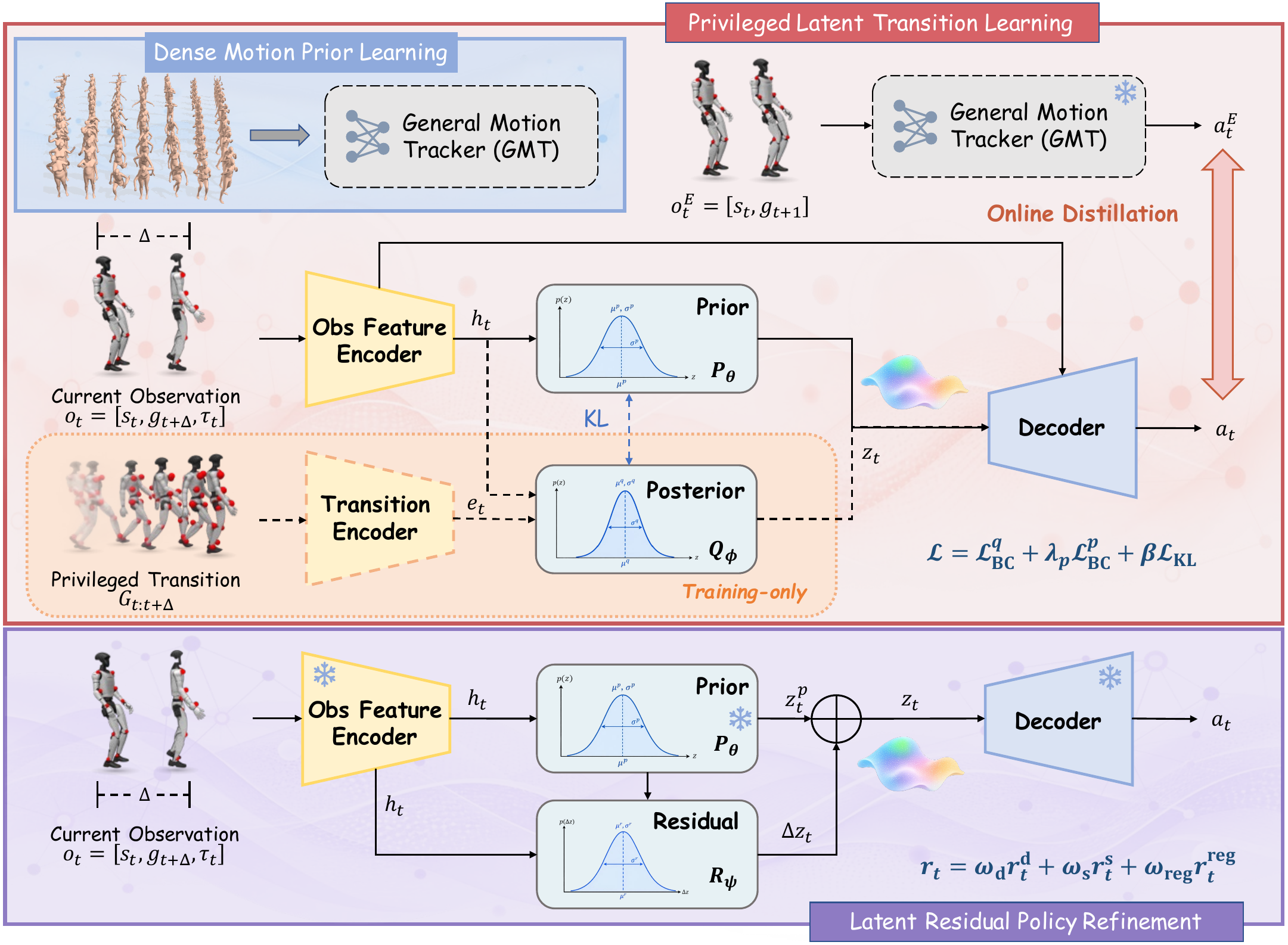}
    \caption{Overview of PLAT. A dense motion-tracking expert provides online action supervision, privileged latent learning distills dense reference transitions into a sparse timed keyframe policy, and latent residual policy refinement adapts the deployed controller while preserving the learned transition prior.}
    \vspace{-5pt}
    \label{fig:stage_method}
\end{figure*}

\subsection{Problem Formulation}
\label{sec:problem_formulation}
We study the task of \emph{Sparse Timed Keyframe Motion Tracking}, where a humanoid robot is required to reach a sequence of sparse future keyframes while maintaining physically plausible and temporally consistent whole-body motions.
Unlike conventional motion tracking, which provides dense frame-by-frame reference motions, our setting exposes the policy only to a sparse future keyframe together with its desired arrival horizon.
The policy must therefore infer the underlying motion transition without access to any intermediate reference frames.

Formally, let $s_t$ denote the current robot state at time step $t$.
A sparse timed command consists of a future target keyframe $g_{t+\Delta}$ together with a normalized horizon
\begin{equation}
\tau_t=\frac{\Delta}{\Delta_{\max}},
\end{equation}
where $\Delta$ denotes the temporal distance between the current frame and the target, and $\Delta_{\max}$ is the maximum command horizon.
The target keyframe is represented in a robot-centric coordinate system, specifying the desired relative pose with respect to the current robot frame.

The policy observation is therefore
\begin{equation}
o_t=\left[s_t,\;g_{t+\Delta},\;\tau_t\right],
\end{equation}
from which the policy predicts the joint position target
\begin{equation}
a_t=\pi(o_t),
\end{equation}
which is executed by the low-level PD controller.

Compared with dense motion tracking, sparse timed keyframe tracking is considerably more challenging because the policy must infer feasible long-horizon motion transitions from highly sparse commands, while the mapping from sparse goals to actions is inherently multi-modal.
Our objective is to learn a sparse policy that accurately reaches timed keyframes while preserving natural and stable whole-body motions.

\subsection{Overview}

The overall framework of PLAT is illustrated in Fig.~\ref{fig:stage_method}. 
To address the missing transition information and the multi-modal nature of sparse timed keyframe tracking, PLAT progressively transfers dense motion knowledge into a deployable sparse policy through three training stages.

The proposed framework consists of three sequential learning phases.
First, a dense motion tracking expert is trained to provide stable motion priors and online action supervision during DAgger~\cite{ross2011reduction}.
Second, a sparse timed keyframe policy is learned through privileged latent transition learning, where dense goal sequences are available only during training to learn a deployable latent transition prior.
Finally, the learned policy is refined through latent residual reinforcement learning, improving sparse timed keyframe tracking while preserving the motion prior acquired through imitation.
At inference, the policy receives only the current robot state, a sparse future keyframe, and its desired arrival time, without requiring dense reference motions or any privileged transition information.

\subsection{Dense Motion Prior Learning}

The first stage learns a general dense motion tracking expert that serves as the motion prior for the subsequent sparse policy learning.
Given the current robot state $s_t$ and the next reference frame $g_{t+1}$, the expert predicts stabilizing whole-body actions $a_t^E=\pi_E(s_t, g_{t+1})$ that accurately track diverse reference motions while maintaining physically feasible behaviors.
Since every control step is guided by dense frame-level supervision, the expert captures rich motion priors covering locomotion, turning, and dynamic whole-body coordination.

The dense expert is trained using a standard frame-by-frame motion tracking objective \cite{luo2025sonic, chen2025gmt, wang2026general}, and remains fixed after convergence.
Rather than being deployed directly for sparse keyframe tracking, it provides a reliable source of motion knowledge for the subsequent learning stages.
Specifically, the expert is later used to guide the sparse policy through online imitation learning, enabling dense motion priors to be progressively transferred to sparse timed keyframe control.

\subsection{Privileged Latent Transition Learning}

While the dense motion expert provides reliable action supervision, directly imitating expert actions from sparse timed commands remains fundamentally challenging. A single future keyframe specifies only the desired destination but provides limited information about how the robot should transition from its current state to the target. As the planning horizon increases, this incomplete transition specification makes it increasingly difficult to infer appropriate intermediate behaviors from sparse observations alone.

To address this ambiguity, we formulate privileged latent transition learning using a \emph{CVAE-style conditional latent variable model}. During training, privileged dense goal sequences are available to infer latent transition representations, while at deployment the policy predicts the latent solely from sparse observations. This formulation enables privileged transition information to be distilled into a deployable latent prior without requiring dense references at inference.

\subsubsection{Privileged Transition Representation}

During training, although the deployed policy observes only the sparse timed command defined in Sec.~\ref{sec:problem_formulation}, the complete reference motion between the current state and the target keyframe remains available.
We therefore construct a privileged goal sequence

\begin{equation}
G_{t:t+\Delta}=\{g_t,g_{t+1},\ldots,g_{t+\Delta}\},
\end{equation}

\noindent which explicitly describes the desired motion transition.
Unlike the sparse timed command, this sequence is available only during training and serves solely as privileged supervision.

The goal sequence is encoded by a GRU-based encoder into a compact transition representation

\begin{equation}
e_t = E_{\phi}(G_{t:t+\Delta}),
\end{equation}

\noindent where $e_t$ summarizes the temporal evolution of the desired transition.
Meanwhile, the sparse observation is processed by the policy backbone,

\begin{equation}
h_t = B_{\theta}(o_t),
\end{equation}

\noindent yielding an observation feature that is shared by both the prior and posterior networks.

Following a CVAE-style formulation, the prior predicts a latent transition distribution conditioned only on the deployable observation,

\begin{equation}
p_{\theta}(z_t|h_t)
=
\mathcal{N}
\!\left(
\mu_t^{p},
\mathrm{diag}\!\left((\sigma_t^{p})^2\right)
\right),
\end{equation}

\noindent where $(\mu_t^{p},\sigma_t^{p})=P_{\theta}(h_t)$.
During training, the posterior additionally conditions on the privileged transition representation,

\begin{equation}
q_{\phi}(z_t|h_t,e_t)
=
\mathcal{N}
\!\left(
\mu_t^{q},
\mathrm{diag}\!\left((\sigma_t^{q})^2\right)
\right),
\end{equation}

\noindent where $(\mu_t^{q},\sigma_t^{q})=Q_{\phi}(h_t,e_t)$.
The posterior latent is sampled using the reparameterization trick,

\begin{equation}
z_t
=
\mu_t^{q}
+
\sigma_t^{q}\odot\epsilon,
\qquad
\epsilon\sim\mathcal{N}(0,I),
\end{equation}

\noindent and the policy predicts the expert action as

\begin{equation}
a_t=\pi_{\theta}(h_t,z_t).
\end{equation}

Unlike conventional goal-conditioned policies that condition only on the target keyframe, the latent variable captures privileged transition information contained in the intermediate reference motion.
By aligning the posterior and prior distributions, the deployable prior learns to infer a transition representation from sparse observations alone, without requiring dense intermediate references during deployment.

\subsubsection{Online DAgger Optimization}

The sparse policy is optimized using online DAgger, where the student interacts with the environment under sparse timed commands while the dense motion expert provides online action supervision.
Compared with offline behavior cloning, DAgger collects expert labels on the state distribution induced by the evolving student policy, thereby reducing compounding errors during long-horizon rollouts.

The posterior and prior policies are supervised by the expert action through


\begin{align}
\mathcal{L}_{\mathrm{BC}}^{q}
&=
\left\|
\pi_{\theta}(h_t,z_t^{q})
-
a_t^{E}
\right\|_2^2,
\\
\mathcal{L}_{\mathrm{BC}}^{p}
&=
\left\|
\pi_{\theta}(h_t,z_t^{p})
-
a_t^{E}
\right\|_2^2,
\end{align}

\noindent where $z_t^{q}\sim q_{\phi}(z_t|h_t,e_t)$ and $z_t^{p}\sim p_{\theta}(z_t|h_t)$.

To transfer privileged transition knowledge into the deployment-time prior, we minimize the KL divergence

\begin{equation}
\mathcal{L}_{\mathrm{KL}}
=
D_{\mathrm{KL}}
\!\left(
q_{\phi}(z_t|h_t,e_t)
\,\Vert\,
p_{\theta}(z_t|h_t)
\right).
\end{equation}

The overall training objective is

\begin{equation}
\mathcal{L}
=
\mathcal{L}_{\mathrm{BC}}^{q}
+
\lambda_p
\mathcal{L}_{\mathrm{BC}}^{p}
+
\beta
\mathcal{L}_{\mathrm{KL}}.
\end{equation}

After training, the posterior network and the goal-sequence encoder are discarded.
The deployed policy retains only the observation encoder, the learned prior, and the actor, enabling sparse timed keyframe tracking using the current robot state, a target keyframe, and its desired arrival time.

\subsection{Latent Residual Policy Refinement}

Although the imitation policy already performs sparse timed keyframe tracking, it is optimized to reproduce expert actions rather than directly maximize the sparse tracking objective.
Directly fine-tuning all policy parameters with PPO often degrades the learned motion prior, leading to unstable behaviors.
Instead, we freeze the observation encoder, prior network, and policy decoder, and optimize only a lightweight residual policy in the latent transition space.

\subsubsection{Latent Residual Policy}

During reinforcement learning, the privileged transition encoder and posterior network are discarded, and the policy operates exactly as at deployment using the deterministic prior latent $z_t^p=\mu_t^p$.
The residual network predicts a latent correction,

\begin{align}
\Delta z_t &= R_{\psi}(h_t,z_t^p),\\
\hat z_t &= z_t^p+\alpha\Delta z_t,\\
a_t &= \pi_\theta(h_t,\hat z_t),
\end{align}

\noindent where $\alpha$ controls the residual magnitude.
Unlike conventional residual RL that directly compensates actions, the proposed formulation performs adaptation in the latent transition space, preserving the learned transition prior while providing sufficient flexibility for task-specific optimization.

\subsubsection{Reinforcement Learning Objective}

The latent residual policy is optimized using PPO \cite{schulman2017proximal}.
The reward combines dense motion tracking objectives, sparse timed keyframe objectives, and regularization terms,

\begin{equation}
r_t
=
w_{\mathrm{dense}}r_t^{\mathrm{dense}}
+
w_{\mathrm{sparse}}r_t^{\mathrm{sparse}}
+
w_{\mathrm{reg}}r_t^{\mathrm{reg}},
\end{equation}

\noindent where the dense tracking reward preserves natural whole-body motions, the sparse reward explicitly optimizes timed keyframe tracking, and the regularization terms encourage smooth and dynamically stable behaviors.

After reinforcement learning, the deployed policy consists only of the observation encoder, the latent transition prior, the latent residual module, and the policy decoder.
The controller therefore requires only the current robot state, a target keyframe, and its desired arrival time, without any dense reference motions or privileged information.
\section{Experiments}

We evaluate PLAT through a series of simulation and real-world experiments designed to answer the following research questions.

\begin{itemize}
    \item \textbf{Q1:} Does PLAT improve sparse timed keyframe tracking compared with existing dense tracking policies?

    \item \textbf{Q2:} Can PLAT maintain accurate and stable motion across different planning horizons, especially for long-horizon sparse tracking?

    \item \textbf{Q3:} How much does each component of PLAT contribute to the final performance, including privileged latent transition learning and latent residual policy refinement?

    \item \textbf{Q4:} Does privileged latent transition learning produce more natural and coherent motion transitions under sparse timed keyframe commands?
\end{itemize}

\subsection{Experimental Setup}

\begin{figure*}[t]
    \centering
    \includegraphics[width=1.0\textwidth]{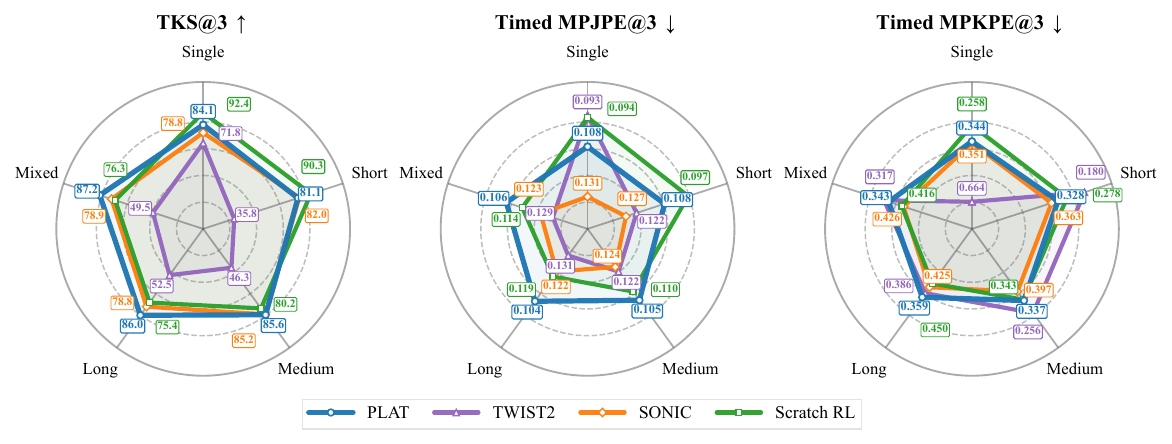}
    \vspace{-15pt}
    \caption{Horizon generalization across sparse timed keyframe intervals. Larger radius denotes better performance for all metrics, and Timed MPJPE@3 and Timed MPKPE@3 use reversed radial scaling for visualization, while the vertex labels show the original values.}
    \label{fig:horizon_generalization}
    \vspace{-5pt}
\end{figure*}

\subsubsection{Evaluation Protocol}

Following the sparse timed keyframe tracking formulation in Sec.~\ref{sec:problem_formulation}, each evaluation episode is constructed from a reference motion by repeatedly sampling a future keyframe together with its normalized arrival horizon.
When the commanded horizon expires, a new future keyframe is issued, resulting in a continuous sequence of sparse timed commands throughout the motion.

To evaluate generalization across different temporal horizons, we consider five evaluation settings:

\begin{itemize}
    \item \textbf{Single}: $\Delta=1$, which degenerates to conventional one-step motion tracking.
    \item \textbf{Short}: $\Delta\in[2,10]$.
    \item \textbf{Medium}: $\Delta\in[11,25]$.
    \item \textbf{Long}: $\Delta\in[26,50]$.
    \item \textbf{Mixed}: $\Delta$ is uniformly sampled from the full horizon range, i.e., $\Delta\in[1,50]$.
\end{itemize}

Unless otherwise specified, all quantitative results are reported under the Mixed setting.

\subsubsection{Evaluation Metrics}

Conventional dense tracking metrics evaluate pose errors at every simulation step and therefore do not faithfully reflect the objective of sparse timed keyframe tracking.
Instead, we evaluate all methods only within the target arrival window, defined as the final three control steps before the commanded horizon expires ($\Delta_t \le 3$).

Based on this protocol, we introduce three task-specific evaluation metrics:

\begin{itemize}
    \item \textbf{Timed Keyframe Success (TKS@3)}: the percentage of commands that successfully reach the target keyframe within the arrival window while satisfying the predefined termination thresholds.

    \item \textbf{Timed MPJPE@3}: the mean per-joint position error evaluated only within the arrival window.

    \item \textbf{Timed MPKPE@3}: the mean per-keypoint position error evaluated only within the arrival window.
\end{itemize}

Unlike conventional dense tracking evaluation, all metrics are computed exclusively around the commanded arrival time, directly measuring the policy's ability to satisfy sparse timed keyframe objectives.

\subsubsection{Baselines}

We compare PLAT with several representative motion tracking policies.

\begin{itemize}

\item \textit{SONIC}~\cite{luo2025sonic} \& \textit{TWIST2}~\cite{ze2025twist2}: representative state-of-the-art dense motion tracking policies. Both methods are designed for frame-by-frame humanoid motion tracking with dense reference motions and serve as strong dense-tracking baselines for comparison with sparse timed keyframe tracking.

\item \textit{Scratch RL}: a general tracking policy trained directly with PPO from scratch.
This policy is also used as the dense motion prior before subsequent DAgger imitation and latent refinement.

\end{itemize}

In addition, we include several variants for ablation:

\begin{itemize}

\item \textit{w/o Privileged Latent}: removing the latent transition model and directly learning a deterministic DAgger policy from sparse observations and expert supervision.

\item \textit{DAgger}: the policy learned through privileged latent transition learning before reinforcement learning refinement.

\item \textit{Direct RL Fine-tune}: initializing PPO from the learned transition policy without residual parameterization, allowing all policy parameters to be updated during reinforcement learning.

\item \textit{Residual Action}: replacing latent residual refinement with conventional action-space residual learning while keeping the same policy initialization.

\item \textit{Residual Latent (PLAT)}: the complete framework with privileged latent transition learning and latent residual policy refinement.

\end{itemize}

\subsection{Main Comparison}
\label{sec:exp_main_comparison}

\begin{table}[!t]
\centering
\caption{Main comparison under the Mixed horizon setting.}
\label{tab:main_comparison_v2}
\resizebox{\columnwidth}{!}{%
\begin{tabular}{lccc}
\toprule
Method & TKS@3 (\%) $\uparrow$ & Timed MPJPE@3 $\downarrow$ & Timed MPKPE@3 $\downarrow$ \\
\midrule
\textit{SONIC} & 78.94 & 0.1231 & 0.4260 \\
\textit{TWIST2} & 49.49 & 0.1291 & \textbf{0.3173} \\
\textit{Scratch RL} & 76.31 & 0.1141 & 0.4160 \\
\textbf{PLAT (Ours)} & \textbf{86.37} & \textbf{0.1065} & 0.3460 \\
\bottomrule
\end{tabular}%
}
\vspace{-8pt}
\end{table}

To answer \textbf{Q1}, we compare PLAT with three representative baselines, \textit{SONIC}, \textit{TWIST2}, and \textit{Scratch RL}, under the Mixed horizon setting.
As shown in Table~\ref{tab:main_comparison_v2}, PLAT achieves the highest TKS@3 together with the lowest Timed MPJPE@3, demonstrating superior sparse timed keyframe tracking performance.
Compared with \textit{Scratch RL}, the improvement indicates that transferring dense motion priors through expert-guided imitation and privileged latent transition learning is substantially more effective than learning motion tracking policies directly from scratch.
Compared with \textit{SONIC}, which is optimized for dense frame-by-frame tracking, PLAT better matches the sparse timed keyframe objective and therefore achieves higher timed success and target-frame accuracy.
Although \textit{TWIST2} obtains the lowest Timed MPKPE@3, its substantially lower TKS@3 indicates frequent tracking failures.

\begin{table*}[t]
\centering
\caption{Ablation study under the Mixed horizon setting.}
\label{tab:ablation_study_v2}
\resizebox{\textwidth}{!}{%
\begin{tabular}{lcccccc}
\toprule
Variant & Privileged Latent & RL Fine-tune & Optimization Space & TKS@3 (\%) $\uparrow$ & Timed MPJPE@3 $\downarrow$ & Timed MPKPE@3 $\downarrow$ \\
\midrule
\textit{w/o Privileged Latent} & \texttimes & \texttimes & -- & 75.57 & 0.1079 & 0.4205 \\
\textit{DAgger} & \checkmark & \texttimes & -- & 83.42 & 0.1080 & 0.3850 \\
\textit{Direct RL Fine-tune} & \checkmark & \checkmark & Full Policy & 12.91 & 0.2205 & 0.5197 \\
\textit{Residual Action} & \checkmark & \checkmark & Action Space & 85.22 & 0.1073 & 0.3574 \\
\textbf{Residual Latent (PLAT)} & \checkmark & \checkmark & Latent Space & \textbf{86.37} & \textbf{0.1065} & \textbf{0.3460} \\
\bottomrule
\end{tabular}%
}
\end{table*}

\begin{figure*}[!t]
    \centering
    \includegraphics[width=1.0\textwidth]{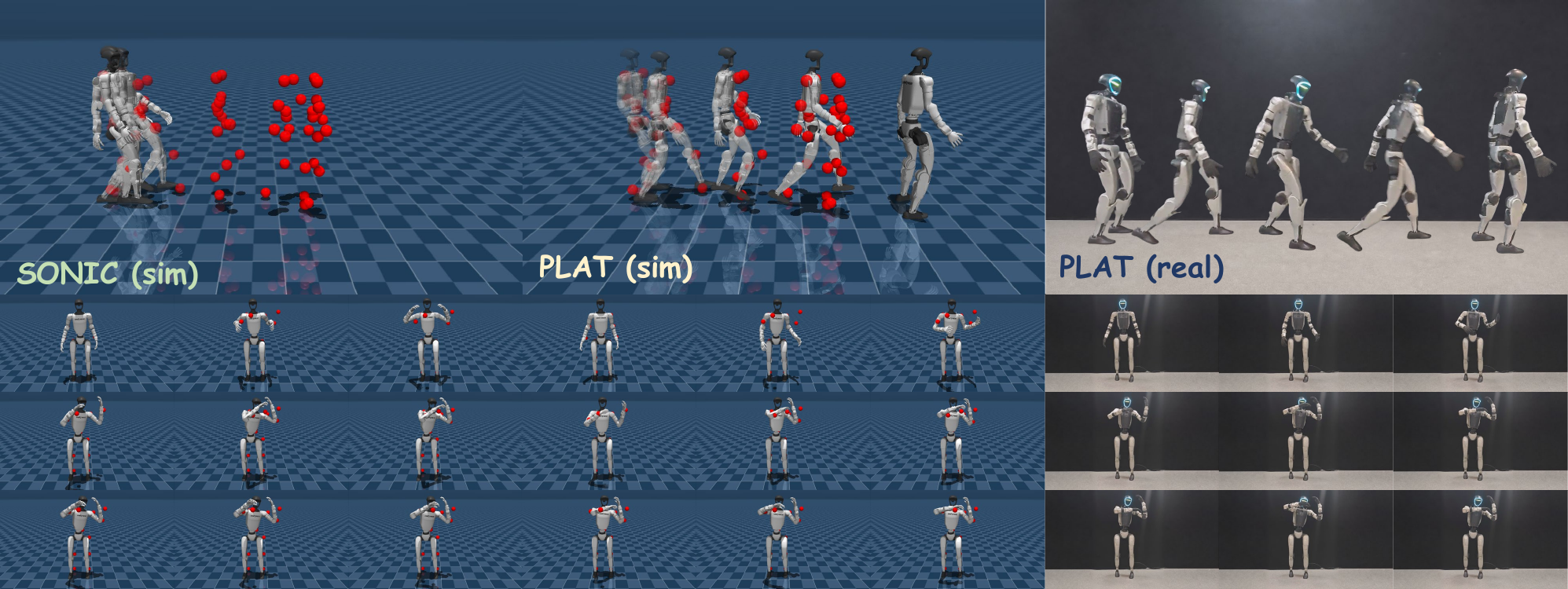}
    \caption{Qualitative comparison of sparse timed keyframe tracking in simulation and the real world. The top row shows a locomotion sequence, while the bottom row shows a dynamic upper-body motion. From left to right, the columns correspond to SONIC in simulation, PLAT in simulation, and PLAT deployed on a Unitree G1 humanoid robot.}
    \label{fig:qualitative_real_robot}
    \vspace{-8pt}
\end{figure*}

\subsection{Horizon Generalization}
\label{sec:exp_horizon_generalization}

To answer \textbf{Q2}, we evaluate the generalization capability of different methods across planning horizons ranging from Single to Long.
The results are shown in Fig.~\ref{fig:horizon_generalization}.

\textit{Scratch RL} performs competitively at short horizons, where the target keyframe remains close to the current state and little transition information is missing.
However, its performance degrades noticeably as the temporal gap between consecutive keyframes increases, indicating limited ability to infer long-horizon motion transitions from sparse commands alone.

Both \textit{SONIC} and \textit{TWIST2} generalize poorly to the proposed sparse timed keyframe setting.
\textit{SONIC} is originally designed to condition on a dense sequence of future reference frames, whereas, for a fair comparison, it is provided only with the commanded target keyframe under our sparse command interface.
Without the future motion context assumed by its original formulation, its performance degrades substantially.
\textit{TWIST2} also exhibits consistently lower success rates together with highly inconsistent pose errors across different horizons, indicating limited robustness under varying temporal gaps.

In contrast, PLAT consistently maintains high success rates and low timed pose errors across all horizon settings.
These results suggest that privileged latent transition learning effectively transfers intermediate transition knowledge into the deployable policy, enabling robust sparse timed keyframe tracking even when intermediate reference motions are unavailable.

\subsection{Ablation Study}
\label{sec:exp_ablation}

To answer \textbf{Q3}, we progressively remove the key components of PLAT to evaluate the contributions of privileged latent transition learning and latent residual policy refinement.
The results are summarized in Table~\ref{tab:ablation_study_v2}.

Comparing \textit{w/o Privileged Latent} with \textit{DAgger} demonstrates the effectiveness of privileged latent transition learning.
Introducing the latent transition model consistently improves both TKS@3 and Timed MPKPE@3, indicating that modeling the intermediate transition between sparse keyframes provides substantially stronger supervision than deterministic imitation from sparse observations alone.
The learned latent representation therefore enables the policy to better infer feasible motion transitions when only sparse timed commands are available at deployment.

The remaining variants investigate different reinforcement learning refinement strategies.
Directly fine-tuning the entire policy with PPO leads to a dramatic performance collapse, suggesting that unconstrained policy optimization quickly destroys the motion prior acquired during imitation learning.
Constraining the update to an action residual substantially improves stability, but still yields inferior performance to latent-space refinement.
Instead of modifying actions directly, PLAT performs residual learning in the learned transition latent space, preserving the underlying motion prior while allowing task-specific adaptation.
As a result, PLAT achieves the highest TKS@3 together with the lowest timed pose errors, demonstrating that latent residual refinement provides a more effective optimization space for sparse timed keyframe tracking.

\subsection{Qualitative and Real-World Visualization}
\label{sec:exp_qualitative_real}

To answer \textbf{Q4}, we visualize representative simulation rollouts and real-robot executions to show how PLAT moves between sparse timed keyframes.
For simulation, we include a \textit{SONIC} tracking rollout as a dense-reference comparison and a PLAT rollout under sparse timed keyframe commands.
The \textit{SONIC} row illustrates the behavior of a dense tracker that receives sparse timed keyframes, whereas the PLAT row shows how sparse commands are connected by the learned latent transition prior.
PLAT produces smoother and more physically plausible sparse-command transitions, reaches the target poses closer to the commanded time, and avoids the overly abrupt, stiff, or unstable recovery behavior observed in \textit{SONIC}.
For hardware validation, we deploy PLAT on a Unitree G1 humanoid robot using the same sparse command representation as in simulation.
The real-world visualization demonstrates that the learned sparse timed keyframe interface and latent transition prior can produce stable whole-body motion beyond simulation.

\section{Discussion and Limitations}

\textit{Sparse Timed Keyframe Motion Tracking} provides a compact control interface by replacing dense frame-level references with sparse future goals.
Compared with conventional motion tracking, the policy must infer appropriate motion transitions from substantially less guidance, making long-horizon execution considerably more challenging.
PLAT addresses this challenge by exploiting dense references only during training: privileged latent transition learning transfers transition information into a deployable latent prior, while latent residual reinforcement learning further improves sparse tracking without disrupting the learned motion prior.

Despite these encouraging results, several limitations remain.
PLAT still relies on dense reference motions and a pre-trained dense tracking expert during training, limiting its applicability when privileged supervision is unavailable.
Moreover, very long planning horizons remain inherently under-specified, making accurate transition inference increasingly difficult.
Future work will investigate richer sparse command representations and methods that further reduce the reliance on privileged supervision.
\section{CONCLUSION}

We presented PLAT, a framework for \textit{Sparse Timed Keyframe Motion Tracking} in humanoid control.
PLAT uses dense motion tracking as a motion prior while removing dense frame-level references from the deployment interface.
By learning goal-sequence privileged latent embeddings through DAgger, the policy transfers privileged dense transition information into a sparse-command prior that operates from only the robot state, a future keyframe, and its desired arrival time.
A latent residual reinforcement learning stage further refines the predicted transition embedding, improving sparse timed execution while preserving the motion prior learned from dense references.
Simulation and real-world experiments on a Unitree G1 humanoid demonstrate that PLAT enables accurate and stable sparse timed keyframe tracking, providing a practical interface between high-level motion planning and low-level whole-body control.

\bibliographystyle{./IEEEtran} 
\bibliography{./ral}

\end{document}